\documentclass[conference]{IEEEtran}

\usepackage{cite}

\usepackage{amsmath}

\usepackage{url}

\usepackage{tabularx}
\usepackage{multirow}
\usepackage{graphicx}
\usepackage{subcaption}
\usepackage{float}
\usepackage{amssymb}
\usepackage{algorithm}
\usepackage{algpseudocode}

\makeatletter
\renewcommand{\ALG@beginalgorithmic}{\footnotesize}
\makeatother

\begin{document}
\bstctlcite{IEEEexample:BSTcontrol}

	\title{Image Embedding and Model Ensembling for Automated Chest X-Ray Interpretation}
	\author{
		\IEEEauthorblockN{Edoardo Giacomello}
		\IEEEauthorblockA{Dipartimento di Elettronica,\\
			Informazione e Bioinformatica\\
			Politecnico di Milano\\
			edoardo.giacomello@polimi.it}
		\and
		\IEEEauthorblockN{Pier Luca Lanzi}
		\IEEEauthorblockA{Dipartimento di Elettronica,\\
			Informazione e Bioinformatica\\
			Politecnico di Milano\\
			pierluca.lanzi@polimi.it}
		\and
		\IEEEauthorblockN{Daniele Loiacono}
		\IEEEauthorblockA{Dipartimento di Elettronica,\\
			Informazione e Bioinformatica\\
			Politecnico di Milano\\
			daniele.loiacono@polimi.it}
		\and
			\IEEEauthorblockN{Luca Nassano}
			\IEEEauthorblockA{Dipartimento di Elettronica,\\
				Informazione e Bioinformatica\\
				Politecnico di Milano\\
				luca.nassano@mail.polimi.it}
		}

	\IEEEoverridecommandlockouts
	\IEEEpubid{\begin{minipage}{\textwidth}\ \\[12pt]
			978-1-7281-1884-0/19/\$31.00 \copyright 2019 IEEE
	\end{minipage}}
	\maketitle

	\begin{abstract}
		Chest X-ray (CXR) is perhaps the most frequently-performed radiological investigation
  globally.
In this work, we present and study several machine learning approaches to
  develop automated CXR diagnostic models.
In particular, we trained several Convolutional Neural Networks (CNN) on
  the CheXpert dataset, a large collection of more than 200k CXR labeled images.
Then, we used the trained CNNs to compute embeddings of the CXR images,
  in order to train two sets of tree-based classifiers from them.
Finally, we described and compared three ensembling strategies to combine
  together the classifiers trained.
Rather than expecting some performance-wise benefits, our goal in this work
  is showing that the above two methodologies, i.e., the extraction of image
  embeddings and models ensembling, can be effective and viable to solve tasks
  that require medical imaging understanding.
Our results in that perspective are encouraging and worthy of further investigation.
 	\end{abstract}

	\IEEEpeerreviewmaketitle

	\section{Introduction}
	\label{sec:introduction}
	The chest X-ray (CXR) is perhaps the most common imaging examination performed for screening,
  diagnostic purposes, and management of many life threatening diseases.
Indeed, automated CXR interpretation could significantly benefit medical practices for tasks like
  the prioritizing of patients in emergency departments or the screening of a large population.
Among the others, machine learning seems a very promising technology to tackle this problem, especially since the recent
  developments on Convolutional Neural Networks (CNNs)\cite{lecun},
  that proved to be very successful when it comes at image understanding and classification~\cite{vgg16}.
Besides the algorithmic and computational improvements, the availability of very large datasets is arguably a
  key element behind the recent successes of these approaches.
Unfortunately, the amount of medical data publicly available is rather limited, both for legal and practical reasons,
  making it very difficult to apply machine learning in this domain.
However, many recent efforts have been made to provide the scientific community with quite large dataset of labeled medical images,
  that allowed the training of machine learning models, able to reach or even to outperform human experts.
In particular, the \emph{CheXpert} dataset\cite{irvin2019chexpert} -- which includes more than 200k CXR
  labeled images -- has been recently made available for a scientific competition on automated CXR
  interpretation \cite{chexpert_competition} that proved how machine learning might be able to achieve outstanding results.

On the other hand, in the medical field we still lack of pre-trained machine learning models that could be
  easily applied to new tasks with a fine-tuning using a rather limited amount of data, as it already happens for
  language\cite{brown2020language} and general-purpose images\cite{imagenet}.
In particular, in the case of medical imaging, making available pre-trained models to extract
  \emph{image embeddings} might be more practical solution.
In fact, \emph{image embeddings} -- low-dimensional representations of the images as a continuous vectors -- can be easily extracted
  using a Convolutional Neural Network (CNN) and used as input to train classifiers based on trees, kernels, Bayesian statistics, etc.
Thus, the advantage of using embeddings lies in retaining the benefit of a CNN trained on a large corpus of images
  while designing a specific classifier for new data and, eventually, for a slightly different problem.
In general, we envision the possibility  of developing a library of embedding models trained by the research community for different
    kind of medical imaging and for different tasks.
Such embedding models and the classifiers trained using them, could be also be combined and mixed together using ensembling strategies.

In this paper, we perform a preliminary investigation of these ideas:
inspired by the work of Pham et al.\cite{pham2019interpreting}, we trained seven different CNNs on the Chexpert dataset and used them
  also to extract image embeddings.
Then, we used these embeddings to train a set of classifiers based on Random Forests~\cite{random_forest} and
  eXtreme Gradient Boosting~\cite{xgboost}.
Finally, we also investigated three different ensembling strategies to combine all the model trained.
Our results are promising and show that image embeddings, as expected, do contain the relevant information to train additional classifiers
  with a performance similar or better than the one achieved using the CNNs in the first place.
We also showed that quite simple ensembling strategies could be used to effectively combine together classifiers, leading to better
  overall performances.

The paper is organized as follows.
In Section~\ref{sec:related} we provide an overview of the most relevant papers on the application of Deep Learning to automate chest
  X-Ray interpretation and in Section~\ref{sec:chexpert} we describe the CheXpert dataset used in this work.
Then, in Section~\ref{sec:methodology} we describe in detail our approach: (i) how we dealt with labels uncertainty,
  (ii) how we exploited the labels dependencies, (iii) the CNN classifiers trained, (iv) how we computed the image embeddings and used them to train additional classifiers, and (v) the ensembling strategies used to combine the trained classifiers.
Thus the experimental design and experimental results are discussed respectively in Section~\ref{sec:expdesign} and Section~\ref{sec:results}.
Finally, we draw our conclusions in Section~\ref{sec:conclusions}.

	\section{Related Work}
	\label{sec:related}
	Along with the availability of large datasets containing chest X-Rays labeled images~\cite{johnson2019mimiccxrjpg,irvin2019chexpert,Wang_2017}, several successful approaches
  based on deep learning and convolutional neural networks have been proposed in the literature.
In their seminal paper~\cite{rajpurkar2017chexnet}, Rajpurkar et al. trained
  a DenseNet-121~\cite{densenet} model on the ChestX-ray14 dataset~\cite{Wang_2017};
their model, dubbed as \emph{CheXNet}, achieved state-of-the-art performance on the classification
of the 14 major thoracic diseases and outperformed expert radiologists on the detection of pneumonia.
In a later work~\cite{rajpurkar2018deep}, Rajpurkar et al. introduced \emph{CheXNeXt} that
  improves the performance of \emph{CheXNet} and achieves a performance similar to expert
  radiologist on 10 thoracic diseases.
Notable works that focus on the ChestX-ray14 dataset are the work of
  Kumar et al.~\cite{kumar2018boosted}, who introduced a cascaded CNNs that can
  diagnose all the 14 thoracic diseases better than baseline, and the work of Lu et
  al.~\cite{Lu2020Multi}, who applied an evolutionary algorithm to search for the most
  suited CNN architecture to solve the classification task.
A different approach was followed by Ye et al.~\cite{ye2020weakly} that introduced
  \emph{Probabilistic-CAM} (PCAM), an extension of CAM~\cite{cam_paper}, to perform the
  localization of thoracic diseases on the ChestX-ray14 dataset in a semi-supervised fashion;
at the same time, the localization model trained can be successfully applied also to solve the
  image classification problem with a performance similar or better than some of the previous
  approaches introduced in the literature, such as \emph{CheXNet}~\cite{Wang_2017}.

More recently, two very large datasets have been released: CheXpert~\cite{irvin2019chexpert} and MIMIC-CXR~\cite{johnson2019mimiccxrjpg}, which  include  respectively 224000 and 350000 images.
Along with the publication of the dataset, Irvin et al.~\cite{irvin2019chexpert} also proposed a
  solution based on a 121-layer DenseNet trained with different approaches to deal with uncertainty
  that is present in the labels of CheXpert dataset.
Their model was able to achieve performance similar or better than expert radiologists on the
  classification of 5 thoracic diseases, selected as the most representative of the dataset.
Instead, Rubin et al.~\cite{rubin2018large} introduced \emph{DualNet}, consisting of two CNNs jointly
  trained on frontal and lateral chest radiographies, included in the very large MIMIC-CXR
  dataset~\cite{johnson2019mimiccxrjpg}.
Their results show that \emph{DualNet} outperforms state-of-the-art classifiers trained separately
  on a single type of image (i.e., either frontal or lateral).

Finally, in a very recent work Pham et al.\cite{pham2019interpreting} trained several
  state-of-the-art CNN on the CheXpert dataset, showing the benefits of exploiting the
  conditional dependencies among the labels in the training as well as of employing an
  ensemble of classifiers with different architectures instead of a single one.

	\section{CheXpert Dataset}
	\label{sec:chexpert}
	From a Machine Learning point of view, CXR interpretation can be modeled as an image classification problem and, thus, requires a
  large dataset that is labeled with quality standards close to the ones provided by expert radiologists.
In this work, we focused on the CheXpert dataset~\cite{chexpert_competition}, that is composed of 223316 CXRs of 65240 patients,
  collected from the Stanford Hospital from October 2002 to July 2017.
The dataset is provided in two different image formats: the high quality format is 16-bit PNG and the low-quality format is 8-bit PNG.
Each image is annotated with a vector of 14 labels, corresponding to major findings in a CXR.
The labels have been extracted from text radiology reports using an automatic rule-based labeler.
In particular, the labeling process consisted of three different phases:
(i) the \emph{Impression} section -- that generally summarizes the key finding of the exam -- of each report is analyzed and a list of mentions is extracted, by matching a list of phrases designed by multiple expert radiologists;
(ii) each mention is assigned to a label according to a level of confidence between \emph{positive}, \emph{negative} and \emph{uncertain}; (iii) each image is encoded as a vector that include one element for each label: positive labels are encoded as 1, negative labels are encoded as 0, uncertain labels are encoded as \emph{u}.
CheXpert dataset includes two kinds of images: frontal and lateral X-Rays. Lateral images are available only for some patients, generally when the diagnosis is uncertain. For this reason, the amount of frontal images is much higher. In addition to the training set, the authors also provide a set of 200 images -- annotated by human experts -- that can be used to assess the performances of the machine learning approaches.
\begin{figure}
	\centering
	\includegraphics[width=1.1\linewidth]{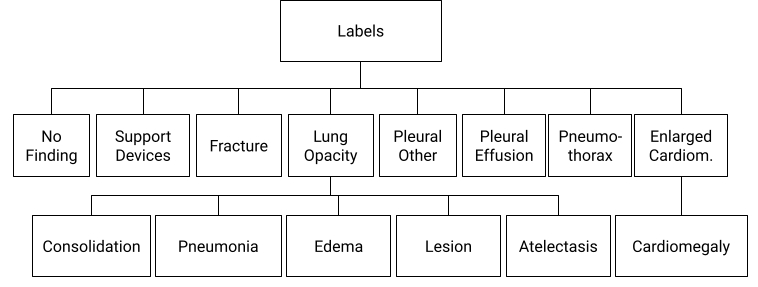}
	\caption{Hierarchical structure of the findings. This is a slightly simplified version of the hierarchy described in \cite{irvin2019chexpert}.}
	\label{fig:label_hirerarchy}
\end{figure}
Table\ref{table:data_distribution} shows the data distribution among the 14 labels included in the dataset.
Following what suggested literature\cite{chexpert_competition,pham2019interpreting}, in this work we focused only
  on five representative findings in CheXpert dataset: Atelectasis, Cardiomegaly, Consolidation, Edema,
	  and Pleural Effusion.
In addition, the 14 findings included in the dataset are not independent but they forms instead an hierarchy as showed in
  Figure~\ref{fig:label_hirerarchy}.
Accordingly, in order to be successful, any machine learning approach applied to CheXpert dataset would need to deal both with
  the uncertainty of the labeling process as well as with the dependency among the labels, that could be exploited to improve
  the performances.

\begin{table}
	\centering
	\begin{tabular}{|l|l|l|l|}
		\hline
		\textbf{Pathology}         &  \hfill \textbf{Positive (\%)}& \hfill \textbf{Uncertain (\%)}  & \hfill \textbf{Negative (\%)}   \\
		\hline
		No Finding &    \hfill 16974 (8.89) &  \hfill 0 (0.0)  & \hfill 174053 (91.11)    \\\hline
		\textbf{Enlarged Card.} &    \hfill 30990 (16.22) &  \hfill 10017 (5.24)  & \hfill 150020 (78.53)    \\\hline
		Cardiomegaly &    \hfill 23385 (12.24) &  \hfill 549 (0.29)  & \hfill 167093 (87.47)    \\\hline
		Lung Opacity &    \hfill 137558 (72.01) &  \hfill 2522 (1.32)  & \hfill 50947 (26.67)    \\\hline
		Lung Lesion &    \hfill 7040 (3.69) &  \hfill 841 (0.44)  & \hfill 183146 (95.87)    \\\hline
		\textbf{Edema} &    \hfill 49675 (26.0) &  \hfill 9450 (4.95)  & \hfill 131902 (69.05)    \\\hline
		\textbf{Consolidation} &    \hfill 16870 (8.83) &  \hfill 19584 (10.25)  & \hfill 154573 (80.92)    \\\hline
		Pneumonia &    \hfill 4675 (2.45) &  \hfill 2984 (1.56)  & \hfill 183368 (95.99)    \\\hline
		\textbf{Atelectasis} &    \hfill 29720 (15.56) &  \hfill 25967 (13.59)  & \hfill 135340 (70.85)    \\\hline
		Pneumothorax &    \hfill 17693 (9.26) &  \hfill 2708 (1.42)  & \hfill 170626 (89.32)    \\\hline
		\textbf{Pleural Effusion} &    \hfill 76899 (40.26) &  \hfill 9578 (5.01)  & \hfill 104550 (54.73)    \\\hline
		Pleural Other &    \hfill 2505 (1.31) &  \hfill 1812 (0.95)  & \hfill 186710 (97.74)    \\\hline
		Fracture &    \hfill 7436 (3.89) &  \hfill 499 (0.26)  & \hfill 183092 (95.85)    \\\hline
		Support Devices &    \hfill 107170 (56.1) &  \hfill 915 (0.48)  & \hfill 82942 (43.42)    \\
		\hline
	\end{tabular}
	\caption{Number of Positive, Uncertain and Negative samples for each finding. In bold the five findings we focused on in this paper.}
	\label{table:data_distribution}
\end{table}

	\section{Methodology}
	\label{sec:methodology}
	In this section, we provide an overview of the machine learning approaches we applied and compared on the CheXpert
  dataset.
In particular, we first describe the approaches we used to deal with label uncertainty and dependency described in the previous section.
Then, we provide an overview of all the machine learning models we compared on the CheXpert dataset:
(i) several models based on convolutional neural networks and (ii) two models based on trees.
Finally, we describe how these models can also be aggregated as an ensembling to compute a better final prediction.

\subsection{Dealing with uncertain labels}
\label{sec:uncertainty}
As described in the previous section, the CheXpert dataset includes a significant number of samples that have been labeled as uncertain. The uncertainty could reflect both an unreliable diagnosis or an ambiguity in the report.
In~\cite{irvin2019chexpert}, Irvin et al. compared different policies to deal with uncertain labels, such as assuming uncertain
  labels as either positive or negative.
On the other hand, Pham et al. \cite{pham2019interpreting} showed that these policies would possibly result in several wrong labels and,
  consequently, could be misleading for training a model.
Accordingly, in this work we followed the \emph{Label-Smoothing Regularization} (LSR) approach introduced by Szegedy et al.~\cite{lsr},
  that allows to exploit the large amount of uncertain labels in CheXpert dataset but prevents the model from becoming over confident on
  uncertain samples.
This is achieved by replacing the uncertain label, \emph{u}, with a random value drawn from a uniform distribution $U(a,b)$
  (with $b>a>0.5$).

\subsection{Exploiting dependencies between labels}
\label{sec:dependencies}
As described in Section~\ref{sec:chexpert}, the labels included in the CheXpert dataset have a hierarchical dependency.
Thus, when training a classification model, such dependencies could be exploited to achieve better performances.
To this purpose, inspired by the work of Pham et al.~\cite{pham2019interpreting}, we employed a \emph{conditional training} approach
  that aims at learning from data the conditional probabilities distribution of labels.
This approach relies an the hierarchical dependency model illustrated in Figure~\ref{fig:label_hirerarchy} and involves a two-steps training
  process.
First, we train our classifiers only with samples that have positive values (i.e., equal to 1) in labels that are not leaves
  in the label hierarchy (i.e., Lung Opacity and Enlarged Cardiomegaly as reported in Figure~\ref{fig:label_hirerarchy}).
Second, we perform an additional training of the classifiers on the whole dataset, to tune their prediction of labels at
  higher level in the hierarchy.
As a result of this two-steps training process, the output of our classifiers can be viewed as the conditional probability that a label is
  positive assuming as positive its \emph{parent} labels (if they exist in the hierarchy).
Accordingly, when conditional training is employed, to predict the unconditional probability of unseen data we simply apply the Bayes
  rule:
we compute the probability for each label as the product of the classifier outputs for that specific label and all the labels above in
  the hierarchy.

\subsection{CXR Classification with CNNs}
\label{ssec:cnn}
In this work we trained and compared seven different convolutional neural networks, that differ  in terms of architectures, topology, and number of parameters. More specifically, the networks considered -- along with the number of parameters -- are DenseNet121 (7M), DenseNet169 (12,5M), DenseNet201 (18M) \cite{densenet},  InceptionResNetV2 (54M) \cite{inceptionresnetv2}, Xception (21M) \cite{xception} , VGG16 (15M) \cite{vgg16} , VGG19 (20M) \cite{vgg16}.
We used different network architectures because, as reported also by other authors~\cite{pham2019interpreting}, each architecture has
  different performances on different labels.
Indeed, since there are no prevailing architectures, an approach based on the aggregation of different models can be beneficial to the final performances as we will later discuss.
In order to use the networks as classifiers, we removed the original dense layer and replaced it with a Global-Average-Pooling (GAP) \cite{gap} layer, followed by a Fully Connected Layer that matches the number of labels in our dataset.

\subsection{CXR Classification with Trees}
\label{ssec:embeddings}
In the recent years, CNNs proved very successful in many image understanding tasks, including CXR interpretation.
Arguably, the main reason behind these successes is the capability of CNNs to learn effective image representations directly
  from data, without the need of design task-specific features.
To investigate better this idea, in this paper we combined CNNs with two well known classifiers based on trees:
  Random Forests~\cite{random_forest} (RF) and eXtreme Gradient Boosting~\cite{xgboost} (XGBoost).
More specifically, we applied the same CNNs trained to classify the CXR images (as described in the Section~\ref{ssec:cnn})
  to extract also a compact image representation, usually called \emph{image embeddings}.
The underlying idea of this process, widely used nowadays in several application domain, is that a CNNs is composed by a
a sequence of convolutional layer typically followed by one or more fully connected layers.
While the convolutional part of a CNN can be basically seen as an universal features extractor, the fully connected layers are
  actually responsible to solve the specific tasks the CNN is applied to (i.e., either a classification or a regression problem).
Thus, we can easily apply a CNN to an image and extract the output of the convolutional part to use it as a large vector of
  features the describe the input image, i.e., an \emph{image embedding}.
In this work, we generated several image embeddings using the different CNNs trained on the CheXpert dataset and used such generated
  dataset to train RF and XGBoost classifiers.

\subsection{Ensembling Strategies}
\label{ssec:ensembling}
 As already mentioned, in a multi-label classification task, like the CheXpert one, it might be difficult to find a single
   classifiers that outperforms the others on each target and it might also happen that for some of the targets no strong classifiers
   are available.
In this setting, we might rely on ensembling strategies that allow to combine several weak classifiers into a stronger one.
In particular, in this paper we investigated three different approaches to combine together multiple classifiers:

\smallskip\noindent
\textbf{Simple Averaging.}
The first approach we used simply consists in averaging the predictions made by the classifiers. If we call $y_{i}(\mathbf{x})$ the
  prediction vector provided as output by the classifier $i$ for the input $(\mathbf{x})$, then the ensemble prediction $\tilde{y}(\mathbf{x})$ computed using $N$ classifiers is:

  \begin{equation}
  	\centering
  	\tilde{y}(\mathbf{x}) = \frac{1}{N}\sum_{i=1}^{N} y_{i}
  	\label{eq:embedding_average}
  \end{equation}

The major drawback of this approach is that it assigns the same weight to all the classifiers, without
  acknowledging that some classifiers may outperform others for specific labels or may be just more confident of their predictions
  for a specific input $(\mathbf{x})$.

\smallskip\noindent
\textbf{Entropy-Weighted Average.}
An alternative approach to simple averaging is to weight each classifier by taking into account their confidence level.
In particular, we developed an heuristic weighting approach based \emph{Entropy}.
In Information Theory, Entropy measures the level of uncertainty of the outcomes of some random variable.
Accordingly, we might model the prediction of each classifier (for a specific label) as as random variable which
  follows a Bernoulli distribution with success probability equal to the actual output of the classifier for that label.
We can thus use the Entropy value of such variable as a measure of the classifier confidence:
 \begin{equation}
 	H_k(p_{k,i}) = - p_{k,i} \log _{2}p_{k,i} - \left(1-p_{k,i} \right) \log _{2}\left( 1-p_{k,i} \right)
 	\label{eq:entropy}
 \end{equation}
 where $p_{k,i}$ is the prediction of the classifier $k$ for label $i$.
As a result, $H_k(p_{i})$ measures the level of uncertainty of the classifier $k$, while $(1-H(p_{i}))$ might be seen as the confidence
  of classifier $k$ about its prediction on label $i$.
In an ensemble of $N$ classifiers that provide a prediction for $L$ labels, for each input we will get
  a \emph{prediction matrix} $P = ( p_{k, i} ) \in \mathbb{R}^{N x L} $.
Applying Equation~\ref{eq:entropy} we can combine for each label $i$the classifiers predictions as follows:
\begin{equation}
 	y_{i} = \sum _{k=1}^{N} \left(1- H_k(p_{k,i}) \right)p_{k,i}
 	\label{eq:weighted_average}
\end{equation}

\smallskip\noindent
\textbf{Stacking.}
The last aggregation approach we investigated involves Stacking \cite{WOLPERT1992241}. This approach combines multiple classification models using a meta-classifier. Specifically, a train set is first used to train the base classifiers, then the predictions of the base models are used as features to train the meta-learner. The pseudo-code for the stacking algorithm is shown in Algorithm \ref{alg:stacking}.

\begin{algorithm}
	\caption{Stacking}\label{euclid}
	\begin{algorithmic}[1]
		\Procedure{StackedGeneralization}{}
		\State $D \gets \{(x_1, y_1), ... ,(x_m, y_m)\}$ \Comment{Training Dataset}
		\State $ \Phi \gets {\Phi_1, \Phi_2, ..., \Phi_T}$ \Comment{Base Classifier}
		\For {$t \gets 1, T$}
			\State $h_y = \Phi_t(D)$
		\EndFor

		\State $D' \gets \emptyset $
		\For {$i = 1, 2, ..., m$}
			\For {$t = 1, 2, ..., T$}
				\State $z_{it} = h_t(x_i)$
			\EndFor
		\State $ D' \gets D'   \cup  \{ ((z_{i1}, z_{i2},..., z_{iT}), y_i) \}$
		\EndFor
		\State {$h' \gets \Phi(D')$ } \Comment{Learn meta-classifier}
		\State  \Return{$h'$}
		\EndProcedure
	\end{algorithmic}
\label{alg:stacking}
\end{algorithm}

	\section{Experimental Design}
	\label{sec:expdesign}

\subsection{Preprocessing}
In this work, for the sake of simplicity, we trained our classifiers only on the frontal images included in the CheXpert dataset,
  as they were present for every patient.
Then, we further split the dataset into a training set (roughly 90\% of samples corresponding to $N=189116$ samples) and a validation set
  (roughly 90\% of samples corresponding to $N=1911$ samples) for tuning the model hyper-parameters.
Thus, we kept the additional set of 202 samples included in CheXpert as test set to assess the final performances of our classifiers.
We pre-processed the data by dropping additional information for each patient such as sex and age.
The uncertain labels were mapped into values sampled from a uniform distribution $U(0.55, 0.85)$, following the LSR approach introduced in
  Section~\ref{sec:methodology}.
Concerning the CXR images included in the dataset, we processed them to reduce as much as possible any noise, such as text or irregular borders, that
  could affect learning performances.
Accordingly, we first resized the images to 256 x 256 pixels and then we applied a template matching algorithm in order to find a region of 224x224 containing a chest template.
Moreover, to match the data shape with those of the network inputs, we converted the images from 1 to 3 channels (RGB), and we scaled their values in the
  range [0,1].
Finally, since the models have been pre-trained on ImageNet, we normalized all the images with respect to that dataset mean and standard deviation.

\subsection{CNN Training}
In our experimental analysis, we trained seven popular CNNs: DenseNet121, DenseNet169, DenseNet201,  InceptionResNetV2, Xception, VGG16, and VGG19.
As described in Section\ref{ssec:cnn}, all these networks have been pre-trained on the ImageNet dataset\cite{imagenet} before training them
  on the CheXpert dataset:
the networks are initialized using the weights provided by the Tensorflow 2.0 Keras module, discarding the classification layer while retaining the
  convolutional layers.
To apply the conditional training approach described in Section\ref{sec:dependencies}, in the first stage, we trained the networks only using the
  samples -- 23526 samples out of 189116 ones -- labeled as positive for all the findings that are not at the bottom of the label hierarchy (see Figure\ref{fig:label_hirerarchy}); then, we froze all the layers except the last fully connected one and we fine-tuned the networks by training them on the whole training set.
In both these training stages, binary cross-entropy was used as loss function and the learning rate was initially set to 1e-4, to be reduced by a
  factor of 10 after each epoch.

Besides training seven different CNNs to classify unlabeled CXRs images, we used them also to compute an image embedding for each sample of the CheXpert
  dataset.
In particular, the embeddings are computing by extracting the output of the Global Average Pooling layer that is before the last fully connected
  layer used to classify the image.
As a result, each trained CNN is able to map a CXR image into a large embedding vector, whose size actually depends on the topology of each CNN as
  illustrated in Table~\ref{table:embeddings}.

\begin{table}
\centering

	\begin{tabular}{|l|l|}
		\hline
		\textbf{Model Name}         &   \textbf{Embedding Shape} \\
		\hline
		DenseNet121 &    (1,1024) \\ \hline
		DenseNet169 &  (1,1664)   \\ \hline
		DenseNet201 &   (1,1920) \\ \hline
		InceptionResNetV2 &   (1,1536) \\ \hline
		Xception &   (1,2048) \\ \hline
		VGG16 &   (1,512) \\ \hline
		VGG19 &   (1,512) \\ \hline
	\end{tabular}

	\caption[Embeddings Shape]{Shape of the CXR image embeddings computed by each CNN.}
	\label{table:embeddings}
\end{table}

\subsection{Training Trees}
Using the image embeddings as an input, we trained two sets of additional classifiers based on trees: for each embedding dataset (computed
  by using each one of the CNNs trained) a RF classifier and an XGBoost classifier were trained.
Concerning the RF classifiers, we performed a grid search to optimize the classifier hyper-parameters based on the performances achieved on the
  validation set.
More specifically we optimized the following hyper-parameters:
(i) \emph{Max Depth}, i.e., the largest tree depth allowed, that regulates the balance between accuracy and overfitting;
(ii) \emph{Min Sample Split}, i.e., the smallest number of samples to allow the split of an internal tree node;
(iii) \emph{Min Sample Leaf}, i.e., the smallest number of samples required for a node to become a leaf of the tree.
Instead we set \emph{Number of Estimators}, i.e., the number of trees in the forest, to 200 (due to computational reasons) and \emph{Max Features},
  i.e., the number of features considered when searching for the best split, was set to the square root of the embedding size.
Such hyper-parameters optimization process was carried out for each RF classifier, corresponding to one of the seven sets of image embeddings computed.
Table \ref{table:rf_parameters} shows the results of the optimization and the final values of the parameters.

 \begin{table}
 	\centering
 		\resizebox{\columnwidth}{!}{
 	\begin{tabular}{|l|l|l|l|l|}
 		\hline
 		\textbf{Model Name} &   \hfil \textbf{Max Depth}&   \hfil \textbf{Min Sample Split}&   \hfil \textbf{Min Sample Leaf}\\
 		\hline
 		DenseNet121 &       \hfil15  &   \hfil2 &   \hfil10  \\\hline
 		DenseNet169 &       \hfil15  &   \hfil2 &   \hfil10  \\\hline
 		DenseNet201 &       \hfil30  &   \hfil10 &   \hfil10  \\\hline
 		InceptionResNetV2&  \hfil30  &   \hfil10 &   \hfil1  \\\hline
 		Xception    &       \hfil30  &   \hfil10 &   \hfil1  \\\hline
 		VGG16       &       \hfil 5  &   \hfil2 &   \hfil 10 \\\hline
 		VGG19       &       \hfil15  &   \hfil50 &   \hfil 1 \\
 		\hline
 	\end{tabular}
 }
 	\caption{Values of the hyper-parameters computed for each RF classifier}
 	\label{table:rf_parameters}
 \end{table}
Concerning XGBoost, instead, we performed an hyper-parameters optimization focused on boosting rounds and the maximum depth of trees.
Our analysis, showed that the best settings for all the classifiers resulted to be using maximum depth equal to 3 and 50 boosting rounds.

\subsection{Stacking}

As described in the Section\ref{ssec:ensembling}, in order to combine the output of the several classifiers trained, we compared three different
  ensembling strategies, including stacking (or stacked generalization).
To apply stacking, it is required to train a sort of meta-classifier that receives as input the output of all the classifiers to combine and computes
  as output the final combined prediction.
In principle, this meta-classifier can be trained used any machine learning method and should possibly not be trained using the same data used to
  train the classifiers combined together in order to avoid possible bias.
In this paper, we used a RF classifier to apply stacking and trained it using the samples in the validation set.
The RF parameters have been empirically set as follows: \emph{Max Tree Depth} was set to 30, \emph{Number of estimators} was set to 1400,
  \emph{Maximum Tree Depth} was set to 30, \emph{Minimum Sample Split} was set to 5, and \emph{Minimum Samples per Leaf} was set to 1.

\subsection{Performance Evaluation}
To assess the performances of our classifiers we computed the \emph{Area Under the Receiving Operating Characteristic} (AUROC),
  i.e., the area under a curve (\emph{Receiving Operating Characteristic}) that is obtained plotting the True Positive Rate (TPR) against the False Positive Rate (FPR) of the classifier, also respectively known as \emph{sensitivity} and \emph{specificity}.
To compute TPR and FPR, the probability predicted by the network needs to be converted into a binary decision, using a threshold between 0 and 1,
  that affects the trade-off between the two metrics.
As a reference, an AUROC value of 0.5 means no discriminative power, while -- in medical field -- a value between 0.7 and 0.8 is considered acceptable,
  a value between 0.8 and 0.9 is considered excellent, and values larger than 0.9 are considered outstanding\cite{auroc}.
Once we set a specific threshold value to use the classifier in practice, it is also possible to plot the confusion matrix of each classifier
  on the test.

	\section{Results}
	\label{sec:results}
	In this section, we present and discuss of our experimental results.
First, we show the results obtained using the classifiers based on CNNs,
 then the results obtained by classifiers based on trees (RF and XGBoost), and
 finally the performances achieved by combining all the best classifiers together.

\subsection{Results of CNNs}
\begin{table}
	\centering
	\resizebox{\columnwidth}{!}{
		\renewcommand{\arraystretch}{1.5}
		\begin{tabular}{|l|l|l|l|l|l|l|}
			\hline
			\textbf{Model} & \textbf{Atelectasis} & \textbf{Cardiomegaly} & \textbf{Consolidation} & \textbf{Edema} & \textbf{P.}~\textbf{Effusion} & \textbf{Mean}  \\
			\hline
			DenseNet121& 			\textbf{0.854} & \textbf{0.800}  &			    0.891 &              0.920 & 			   0.917 & \textbf{0.876}            \\ \hline
			DenseNet169&						 0.850 & 				0.795  & 			   0.882 & \textbf{0.936} & 			0.915 & \textbf{0.876}            \\ \hline
			DenseNet201& 						 0.834 & 				0.791  & 			   0.881 &              0.917 & 			  0.925 & 			   0.870            \\ \hline
			InceptionResNetV2&			 	 0.816 & 				0.784  & 			   0.897 &              0.925 & 			  0.919 & 			   0.869            \\ \hline
			Xception& 								  0.841 & 			     0.770  & 				0.880 &              0.909 & 			   0.924 & 				 0.865            \\ \hline
			VGG16& 								       0.843 & 				 0.772  & 				0.898 &              0.932 & 			   0.919 & 				  0.873            \\ \hline
			VGG19& 									   0.843 &				 0.769  & \textbf{0.900} &              0.927 & \textbf{0.933} & 			   0.874            \\ \hline

		\end{tabular}
	}
	\caption{Performances of the CNNs trained. The name of different CNN architecture is reported in the \texttt{Model} column.
  The performance is computed as the AUROC achieved on test set. The \texttt{Mean} column shows the average performance on all the five findings. We reported in bold the best performance for each finding and overall.}
	\label{table:cnn_models}
\end{table}
Table \ref{table:cnn_models} shows the results achieved by each CNN trained for each one of the main
   five findings considered: Atelectasis, Cardiomegaly, Consolidation, Edema,
 	  and Pleural Effusion.
The results show that all the CNNs achieve a similar average AUROC over the five findings
  and they also have very similar performances on each single finding:
all the networks achieves outstanding performances on identifying Edema and Pleural Effusion, while they struggle at detecting Cardiomegaly.
On the other hand, the results show that, as expected, there is no a single CNN that outperforms the
  others consistently for all the five findings: as an example, VGG19 achieves the best performance on Consolidation and Pleural Effusion but also the worst performance on Cardiomegaly.
For this reason we applied three different ensembling strategies, described in Section \ref{ssec:ensembling}, to combine all the seven CNNs trained.

\begin{table}
	\centering
	\resizebox{\columnwidth}{!}{
		\renewcommand{\arraystretch}{1.5}
		\begin{tabular}{|l|l|l|l|l|l|l|}
			\hline
			\textbf{Method} & \textbf{Atelectasis} & \textbf{Cardiomegaly} & \textbf{Consolidation} & \textbf{Edema} & \textbf{P.}~\textbf{Effusion} & \textbf{Mean}  \\
			\hline
			Best CNN& 0.854 &0.800  & 0.900 & \textbf{0.936} & 0.933 & 0.885\\\hline

			Simple Average& 0.854 & \textbf{0.811}  & 0.908 & \textbf{0.936} & 0.933 & 0.888\\\hline

			Entropy Weighted Avg. & \textbf{0.856} & \textbf{0.811}  & \textbf{0.912} & \textbf{0.936} & 0.930 & \textbf{0.889}\\\hline

			Stacking & 0.842 & 0.797  & 0.871 & 0.921 & \textbf{0.937} & 0.873\\\hline

		\end{tabular}
	}

	\caption{Comparison of the performances of the best CNN classifier for each finding (reported as \texttt{Best CNN}) and the three ensembling strategies
    considered.
  The performance is computed as the AUROC achieved on test set.
  We reported in bold the best performance for each finding and overall.}

	\label{table:cnn_ensemble}
\end{table}
Table \ref{table:cnn_ensemble} shows the results of such different ensembling strategies on each of the five findings along with the performance of the best CNNs for that specific finding.
The results show that, except for the stacking approach, the ensembling strategies based on averaging
  allow to achieve overall better performances, exploiting the differences among the CNNs.
In particular the strategy based on entropy-weighted average resulted to be the best achieving a
  slightly better performance overall and for all findings except for Pleural Effusion.
Interestingly, Pleural Effusion is the only target that benefit of a stacking approach,
    perhaps suggesting that it requires a more sophisticated ensembling strategy.

\subsection{Results of RF}
\begin{table}
	\centering
	\resizebox{\columnwidth}{!}{
		\renewcommand{\arraystretch}{1.5}
		\begin{tabular}{|l|l|l|l|l|l|l|}
			\hline
			\textbf{Model} & \textbf{Atelectasis} & \textbf{Cardiomegaly} & \textbf{Consolidation} & \textbf{Edema} & \textbf{P.}~\textbf{Effusion} & \textbf{Mean}  \\
			\hline
			RF+DenseNet121 & 0.851 & \textit{0.818}  & 0.885 & 0.915 & \textbf{\textit{0.945}} & \textit{0.883}\\ \hline
			RF+DenseNet169& \textit{0.855} & \textit{0.814}  & \textit{0.893} & \textbf{0.922} & \textit{0.933} & \textbf{\textit{0.884}}\\ \hline
			RF+DenseNet201& \textit{0.863} & \textit{0.814}  & 0.878 & \textbf{\textit{0.922}} & \textit{0.936} & \textit{0.882}\\ \hline
			RF+InceptionResNetV2& \textit{0.830} & 0.779  & \textit{0.898} & 0.918 & \textit{0.933} & \textit{0.872}\\ \hline
			RF+Xception& 0.831 & \textit{0.810}  & \textit{0.907} & \textit{0.913} & \textit{0.932} & \textit{0.879}\\ \hline
			RF+VGG16& \textit{0.858} & \textbf{\textit{0.822}}  & \textbf{\textit{0.913}} & 0.886 & 0.917 & \textit{0.879}\\ \hline
			RF+VGG19& \textbf{\textit{0.873}} & \textit{0.798}  & 0.895 & 0.892 & 0.917 & \textit{0.875}\\ \hline
		\end{tabular}
	}
  \caption{Performances of the RF classifiers. In the \texttt{Model} column we reported the name of the CNN used to generated the image embeddings
    the RF classifier was trained from.  The performance is computed as the AUROC achieved on test set. We reported in bold the best performance for
    each column and in italic the performances that are better than the corresponding one achieved by the CNN.}
	\label{table:results_randomforest}
\end{table}
The second set of experiments we performed consisted of training RF classifiers using the image embeddings extracted by the previously trained CNNs.
Table \ref{table:results_randomforest} shows the performances achieved by each RF classifiers trained (in Table \ref{table:results_randomforest} we
  reported in \texttt{Model} column the name of the CNN used to generate the image embeddings).
The results show that the RF classifiers achieve in general a better performance than the CNN used to generate the image embeddings (reported in
  italic in Table \ref{table:results_randomforest}) and this is always the case if we consider the mean performance.
This confirms, as expected, that embeddings do actually encode all the relevant information to discriminate the findings and, more interestingly,
  that Random Forests are well suited to replace the last fully connected layer used in the network to perform classification.
In addition, as for the classifiers based on CNNs, also in this case there is not a single classifier that outperforms consistently all the others,
  suggesting that there is the possibility of exploiting ensembling strategies to improve the performances.
  \begin{table}
  	\centering
  	\resizebox{\columnwidth}{!}{
  		\renewcommand{\arraystretch}{1.5}
  		\begin{tabular}{|l|l|l|l|l|l|l|}
  			\hline
  			\textbf{Method} & \textbf{Atelectasis} & \textbf{Cardiomegaly} & \textbf{Consolidation} & \textbf{Edema} & \textbf{P.}~\textbf{Effusion} & \textbf{Mean}  \\
  			\hline
  			Best RF& 0.855 & 0.814  & 0.893 & 0.922 & 0.933 & 0.884\\\hline
  			Simple Average& 0.859 & \textbf{0.828}  & \textbf{0.918} & \textbf{0.921} & \textbf{0.940} & 0.893\\\hline
  			Entropy Weighted Avg.& \textbf{0.872} & 0.826  & \textbf{0.918} & 0.916 & 0.936 & \textbf{0.897}\\\hline
  			Stacking& 0.840 & 0.761  & 0.883 & 0.908 & 0.937 & 0.866\\\hline

  		\end{tabular}
  	}
    \caption{Comparison of the performances of the best RF classifier for each finding (reported as \texttt{Best RF}) and the three ensembling strategies
      considered.
    The performance is computed as the AUROC achieved on test set.
    We reported in bold the best performance for each finding and overall.}
  	\label{table:results_randomforest_ensemble}
  \end{table}
Indeed, Table \ref{table:results_randomforest_ensemble} compares the performances of the three ensembling strategies applied to RF classifiers along
  with the best performance achieved by a single RF classifier -- which is not the same one for each finding.
The results show that the ensembling strategies always outperform the single best RF classifiers and the entropy-weighted average achieved the best
  mean performance overall but simple average performs better in most of the findings.
Despite the differences are very small, this can be easily explained by noting that entropy-weighted average outperforms simple average on identifying
  Atelectasis, where there is a single RF classifier much better than all the others (see Table \ref{table:results_randomforest}):
in this case, the entropy-weighted average can better exploit the most confident classifier weighting its prediction more than the others.
Instead, in this case the stacking approach does not provide any improvement with respect to the most simple ensembling strategies.

\subsection{Results of XGBoost}
\begin{table}
	\centering
	\resizebox{\columnwidth}{!}{
		\renewcommand{\arraystretch}{1.5}
		\begin{tabular}{|l|l|l|l|l|l|l|}
			\hline
			\textbf{Model} & \textbf{Atelectasis} & \textbf{Cardiomegaly} & \textbf{Consolidation} & \textbf{Edema} & \textbf{P.}~\textbf{Effusion} & \textbf{Mean}  \\
			\hline
			XGB+DenseNet121& 0.803 & \textit{0.837}  & 0.837 & \textbf{\textit{0.944}} & \textit{0.935} & 0.871\\ \hline
			XGB+DenseNet169& 0.812 & \textit{0.829}  & 0.866 & 0.916 & \textit{0.923} & 0.869\\ \hline
			XGB+DenseNet201& \textbf{0.824} & \textbf{\textit{0.867}}  & 0.876 & \textit{0.920} & \textbf{\textit{0.938}} & \textbf{\textit{0.885}}\\ \hline
			XGB+InceptionResNetV2& \textit{0.820} & \textit{0.792}  &\textbf{\textit{0.911}} & 0.911 & \textit{0.922} & \textit{0.871}\\ \hline
			XGB+Xception& 0.803 & \textit{0.804}  & \textit{0.901} & 0.899 & 0.923 & \textit{0.866}\\ \hline
			XGB+VGG16& 0.800 & \textit{0.840} & 0.849 & 0.922 & \textit{0.927} & 0.868\\ \hline
			XGB+VGG19& 0.811 & \textit{0.819}  & 0.832 & 0.921 & 0.922 & 0.861\\ \hline
		\end{tabular}
	}
    \caption{Performances of the XGBoost classifiers. In the \texttt{Model} column we reported the name of the CNN used to generated the image embeddings
      the XGBoost classifier was trained from.  The performance is computed as the AUROC achieved on test set. We reported in bold the best performance for
      each column and in italic the performances that are better than the corresponding one achieved by the CNN.}
	\label{table:results_xgboost}
\end{table}
A set of experiments, very similar to the previous ones, have been performed for XGBoost.
Table \ref{table:results_xgboost} shows the performance of the XGBoost classifiers trained with the image embeddings generated
  by CNNs.
The results shows that, except for the \texttt{XGB+DenseNet201}, the mean performances achieved with XGBoost are slightly worse or very close to
  the ones achieved by the corresponding CNNs.
The only notable exception are the performances achieved on Cardiomegaly, where the XGBoost classifiers outperforms both CNNs and the RF classifiers.
This suggest that for that specific findings, the boosting mechanism has a larger impact on the performances.
Similarly to what previously done, we applied ensembling strategies also to XGBoost classifiers.
\begin{table}
	\centering
	\resizebox{\columnwidth}{!}{
		\renewcommand{\arraystretch}{1.5}
		\begin{tabular}{|l|l|l|l|l|l|l|}
			\hline
			\textbf{Method} & \textbf{Atelectasis} & \textbf{Cardiomegaly} & \textbf{Consolidation} & \textbf{Edema} & \textbf{P.}~\textbf{Effusion} & \textbf{Mean}  \\
			\hline
			Best XGBoost         & 0.824           & \textbf{0.867}  & \textbf{0.911} & \textbf{0.944} & 0.938          & 0.885\\\hline
			Simple Average       & 0.829           & 0.863           & 0.902          & 0.933          & 0.939          & 0.893\\\hline
			Entropy Weighted Avg.& \textbf{0.839}  & 0.864           & 0.899          & 0.936          & \textbf{0.940} & \textbf{0.896}\\\hline
		\end{tabular}
	}
  \caption{Comparison of the performances of the best XGBoost classifier for each finding (reported as \texttt{Best XGB}) and the two ensembling
    strategies considered.
  The performance is computed as the AUROC achieved on test set.
  We reported in bold the best performance for each finding and overall.}
	\label{table:results_xgboost_ensemble}
\end{table}
Table \ref{table:results_xgboost_ensemble} shows the performances achieved with the different ensembling strategies along with the one of the best
  XGBoost classifier -- the performances of stacking strategy were worse than the one of simple and entropy-weighted averages and have not been
  reported in the Table.
The results show that, despite ensembling strategies are outperformed on some of the findings by the best single XGBoost classifier, overall
  they achieve a better mean performance than single XGBoost classifiers.
Also, the results show that entropy-weighted average is slightly better than simple average to combine classifiers, consistently with what we
  previously found.

\subsection{Final Results}
Finally, we combined together the three ensembles of classifiers presented so far, by applying the entropy-weighted average approach that
  resulted to be the most reliable one.
\begin{table}
	\centering
	\resizebox{\columnwidth}{!}{
		\renewcommand{\arraystretch}{1.5}
		\begin{tabular}{|l|l|l|l|l|l|l|}
			\hline
			\textbf{Method} & \textbf{Atelectasis} & \textbf{Cardiomegaly} & \textbf{Consolidation} & \textbf{Edema} & \textbf{P.}~\textbf{Effusion} & \textbf{Mean}  \\
			\hline
			DenseNet121& 0.854 & 0.800  & 0.891 & 0.920 & 0.917 & 0.876\\\hline
			RF+DenseNet169& 0.855 & 0.814  & 0.893 & 0.922 & 0.933 & 0.884\\\hline
			XGB+DenseNet201& 0.824 & \textbf{0.867}  & 0.876 & 0.920 & 0.938 & 0.885\\ \hline
			CNN Ensemble& 0.855 & 0.811  & 0.912 & 0.936 & 0.930 & 0.889\\\hline
			RF Ensemble& \textbf{0.872} & 0.826  & \textbf{0.918} & 0.916 & 0.936 & 0.897\\\hline
			XGB Ensemble& 0.839 & 0.864  & 0.899 & \textbf{0.936} & \textbf{0.940} & 0.896\\\hline
			Final Ensemble& 0.860 & 0.860  & 0.917 & 0.934 & 0.939 & \textbf{0.902}\\\hline		\end{tabular}
	}
  \caption{Summary of the performances achieved by the best classifiers and ensembles developed in this work, along with the performances of the final
    ensemble.
  The performance is computed as the AUROC achieved on test set.
  We reported in bold the best performance for each finding and overall.}
	\label{table:results_final}
\end{table}
Table \ref{table:results_final} shows the performances of this final ensemble (dubbed \texttt{Final Ensemble} in the Table) along with the performances
  of the best classifiers trained for each method (CNN, Random Forest and XGBoost) and with the performances of the three ensembles previously presented.
As expected, the results show that such final ensemble allows to combine the benefit of the different approaches and achieves the best overall
  performance (AUROC value of 0.902) with respect to the previously discussed solution.

\begin{figure*}
	\centering
\begin{subfigure}{0.185\textwidth}
	\begin{tabular}{c}
		\includegraphics[width=\columnwidth]{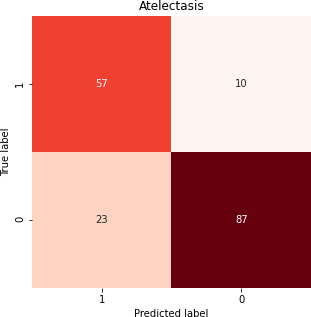} \\
		\footnotesize{Uncertain: 25}
	\end{tabular}
\end{subfigure}
\begin{subfigure}{0.185\textwidth}
	\begin{tabular}{c}
		 \includegraphics[width=\columnwidth]{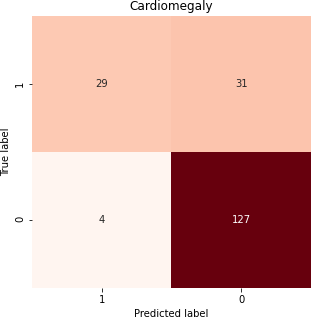} \\
	    \footnotesize{Uncertain: 11}
	\end{tabular}
\end{subfigure}
\begin{subfigure}{0.185\textwidth}
	\begin{tabular}{c}
		\includegraphics[width=\columnwidth]{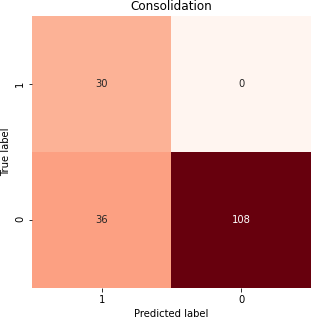} \\
		\footnotesize{Uncertain: 28}
	\end{tabular}
\end{subfigure}
\begin{subfigure}{0.185\textwidth}
	\begin{tabular}{c}
		\includegraphics[width=\columnwidth]{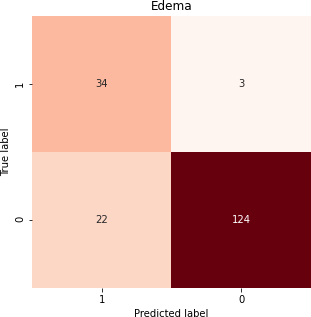} \\
		\footnotesize{Uncertain: 19}
	\end{tabular}
\end{subfigure}
\begin{subfigure}{0.185\textwidth}
	\begin{tabular}{c}
		\includegraphics[width=\columnwidth]{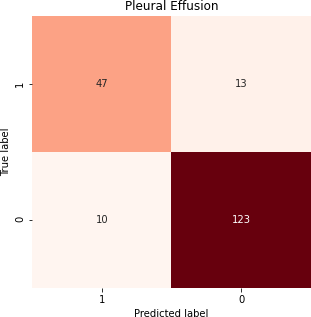} \\
		\footnotesize{Uncertain: 9}
	\end{tabular}
\end{subfigure}
\caption{Confusion matrices achieved classifying the 209 samples in the test set with the final ensemble model.}
\label{fig:figure_5.14}
\end{figure*}

Finally, we wanted also to provide an insight of the classification results that can be achieved in practice when using the
  final ensemble just presented.
To this purpose we set, for each of the five findings, a threshold to discriminate among positive and negative when labeling unseen data.
In particular, we set as a threshold the average model output on the validation set.
In addition, to avoid mistakes on sample too close to the discriminant threshold, we labeled as \emph{uncertain} the output within
  a specific range.
Figure \ref{fig:figure_5.14} shows the resulting confusion matrices for each finding computed on the test set.
This provides a more immediate understanding of the final performances with respect to the AUROC values previously discussed.
In particular we can notice that the number of uncertain samples are approximately among the 5\% and the 15\% of the total, which
  seems to be a reasonable amount of samples that would require a scrutiny instead of being labeled automatically.
The results also shows that, in general, more false positives than false negative are generated by the model.
Despite this was not achieved intentionally, it seems a desirable outcome for a diagnostic model -- the only exception
  being the results on Cardiomegaly, suggesting that perhaps a less conservative threshold might be chosen.

  \section{Conclusions}
  \label{sec:conclusions}
  Chest X-ray (CXR) is a standard diagnostic tool, widely used in the clinical practice.
Thus, reliable methods to automate CXR interpretation could benefit the work-flow of
  doctors.
Machine learning is a promising technology to solve this task, as proved by the recent
  \emph{CheXpert} competition\cite{chexpert_competition} that made available a large dataset
   which includes more than 200k CXR labeled images.
The goal of this paper is twofold:
(i) investigate whether the embeddings of the CXR images extracted from CNNs might be used to train novel classifiers from scratch;
(ii) study the benefit of model ensembling and comparing different ensembling strategies.
To this purposes, we trained on the Chexpert dataset several CNNs: DenseNet121, DenseNet169, DenseNet201,  InceptionResNetV2, Xception, VGG16,
 and VGG19.
Then, we used them to extract embeddings of the images in the dataset and trained from them two sets of classifiers with Random Forest
  and with eXtreme Gradient Boosting.
Finally, we applied and compared three different ensembling strategies to combine all the models trained.
Our results, although preliminary, are promising:
the image embeddings do retain the enough information to train effective classifiers based on trees, achieving a final performance that
  is often even better than the one achieved by the CNN model used to extract the embeddings in the first place.
Also, model ensembling resulted quite useful to combine classifiers, especially when none of the classifier is better than the others on
  all the labels.
More specifically, our results showed that entropy-weighted averaging of the models predictions allow to achieve an overall better performance, by
  weighting more the most confident classifiers for each label.

However, further studies are needed to confirm our findings.
In particular, in future works we plan to exploit the embedding models to train classifiers on other (public and private) CXR datasets
  to solve similar classification tasks, in order to compare the performances with the ones of classifiers trained from scratch.
In addition, we will also investigate the application of convolutional autoencoder to extract more general and unbiased image embeddings, that
  could be possibly be used for many different kinds of tasks (e.g., prognosis prediction, target localization, etc.).

	\bibliographystyle{ieeetran}

\end{document}